\documentclass[letterpaper, 10 pt, conference]{ieeeconf}  
\IEEEoverridecommandlockouts                               
\usepackage{graphicx}                       
\usepackage{graphics}                       
\usepackage{epsfig}                         
\usepackage[tight,footnotesize]{subfigure}  

\usepackage{amssymb,amsmath}
\usepackage{mdwmath}
\usepackage{commath}   
\usepackage{eqparbox}
\usepackage{mathtools}
\usepackage[utf8]{inputenc} 
\usepackage[english]{babel}
\usepackage{amsmath,amsfonts,amssymb}

\newcommand{\et}{{\em et al.\ }}

\newcommand{\beq}{\begin{equation}}
\newcommand{\eeq}{\end{equation}}
\newcommand{\bear}{\begin{eqnarray}}
\newcommand{\bears}{\begin{eqnarray*}}
\newcommand{\eear}{\end{eqnarray}}
\newcommand{\eears}{\end{eqnarray*}}
\newcommand{\bdm}{\begin{displaymath}}
\newcommand{\edm}{\end{displaymath}}
\newcommand{\lba}{\left[\begin{array}}
\newcommand{\ear}{\end{array}\right]}

\newcommand{\degree}{^\circ}     

\usepackage{stfloats}                       
\usepackage{url} 
\usepackage{cite}                           
\usepackage[T1]{fontenc} 
\usepackage{tabularx}
\usepackage{multirow}
\usepackage[table]{xcolor}
\usepackage{longtable}
\usepackage{booktabs} 			
\usepackage{xcolor,colortbl}    

\title{\LARGE \bf Plastic Waste is Exponentially Filling our Oceans, \\but where are the Robots?}
\author{Juan Rojas\\
}
\begin{document}
\maketitle
\thispagestyle{empty}
\pagestyle{empty}
\bstctlcite{IEEEexample:BSTcontrol}
\begin{abstract}
Plastic waste is filling our oceans at an exponential rate. The situation is catastrophic and has now garnered worldwide attention. 
Despite the catastrophic conditions, little to no robotics research is conducted in the identification, collection, sorting, and removal of plastic waste from oceans and rivers and at the macro- and micro-scale. Only a scarce amount of individual efforts can be found from private sources. 
This paper presents a cursory view of the current plastic water waste catastrophe, associated robot research, and other efforts currently underway to address the issue. 
As well as the call that as a community, we must wait no longer to address the problem. Surely there is much potential for robots to help meet the challenges posed by the enormity of this problem.
\end{abstract}
\section{INTRODUCTION}\label{sec:Intro}
Plastic waste in the world waterways has become a global catastrophe. Around \textit{five trillion} pieces of plastic weighing around 322 million tonnes are currently littering our oceans \cite{2014PLoS-Eriksen-5TrillPlastic} and that number is growing with incredible speed by the second.

Despite such catastrophic conditions, very little to none robotics research is being conducted to to find concrete solutions to the identification, collection, sorting, and removal of plastic waste from oceans and rivers and at the macro- and micro-scale. Only a scarce amount of individual efforts can be found (and they originate from a handful of companies and international foundations). 

This paper wishes to question, if the individual technologies required to tackle this monstrous problem exist, why is there no significant research efforts being conducted in trying to address their effectiveness and synergistic effects to solve this problem in the literature? Why are there such few commercial or humanitarian efforts towards this cause? 

In this paper we present a cursory view of the current plastic water waste catastrophe, associated robot research, and other efforts currently underway to address the issue. The goal of the paper is to spur interest, perhaps even indignation within the robotics community, to further study how we can meet the great challenges posed by this situation. We believe that robotics has tremendous potential to offer realistic solutions to the catastrophe of plastic waste in our waters.

The rest of the paper is organized as a survey: Sec. \ref{sec:plastic_problem}, overviews the plastic waste situation in waterways, Sec. \ref{sec:robot_trash_collection}, overviews robotics research focused on trash collection, Sec. \ref{sec:surface_collection}, overviews current efforts to clean oceans and rivers, and finally Sec. \ref{sec:conclusion} offers some summarizing remarks and makes a call to action.
\section{Plastic Problem Overview} \label{sec:plastic_problem}
In this section, we first present a brief overview of concepts relating plastic biodegradation in ocean waters to highlight the problem they pose to maritime life as well as human health. We then continue another section with a number of statistics that detail the current conditions of plastic in our world's oceans. A third subsection narrows the focus to the dynamics of garbage gyres that exist around the world, and in particular, to the largest one in our planet. Finally, we will discuss the damage caused by plastic waste to maritime life. The goal of this section is to provide context for the type of robot solutions necessary to address existing challenges. 
\subsection{Maritime Plastic Pollution, where are we?}
In 2017, Lebreton \et  \cite{2017Nature-Lebreton-RiverEmissionsWorldOceans} quantified marine plastic sources. A global model of plastic inputs from rivers into oceans was presented. With this study along with statistics from the United Nations (UNs) Cleanseas and World Environemnt day portals \cite{2018Cleanseas-UN,2018WorldEnvironmentDay-UN}, we present an overview of the plastic waste pollution catastrophe in our oceans. 

It is estimated that around \textit{five trillion} pieces of plastic (or 322 million tonnes of plastic waste) litter our oceans \cite{2014PLoS-Eriksen-5TrillPlastic}. Each year, rivers, especially those of Asia (20 rivers or 67\% of the total), spew around 2 million tonnes of plastic waste. 60-90\% of marine litter stems from plastic polymers, which come in all sizes, and do not sink once they encounter the sea. During the wearing process they fragment, spread, and take 500 to 1000 years to decompose.
\subsection{Plastic Fragmentation and Biodegradation in Ocean Waters}
\begin{figure}[t]
  \centering
  \includegraphics[width=\linewidth]{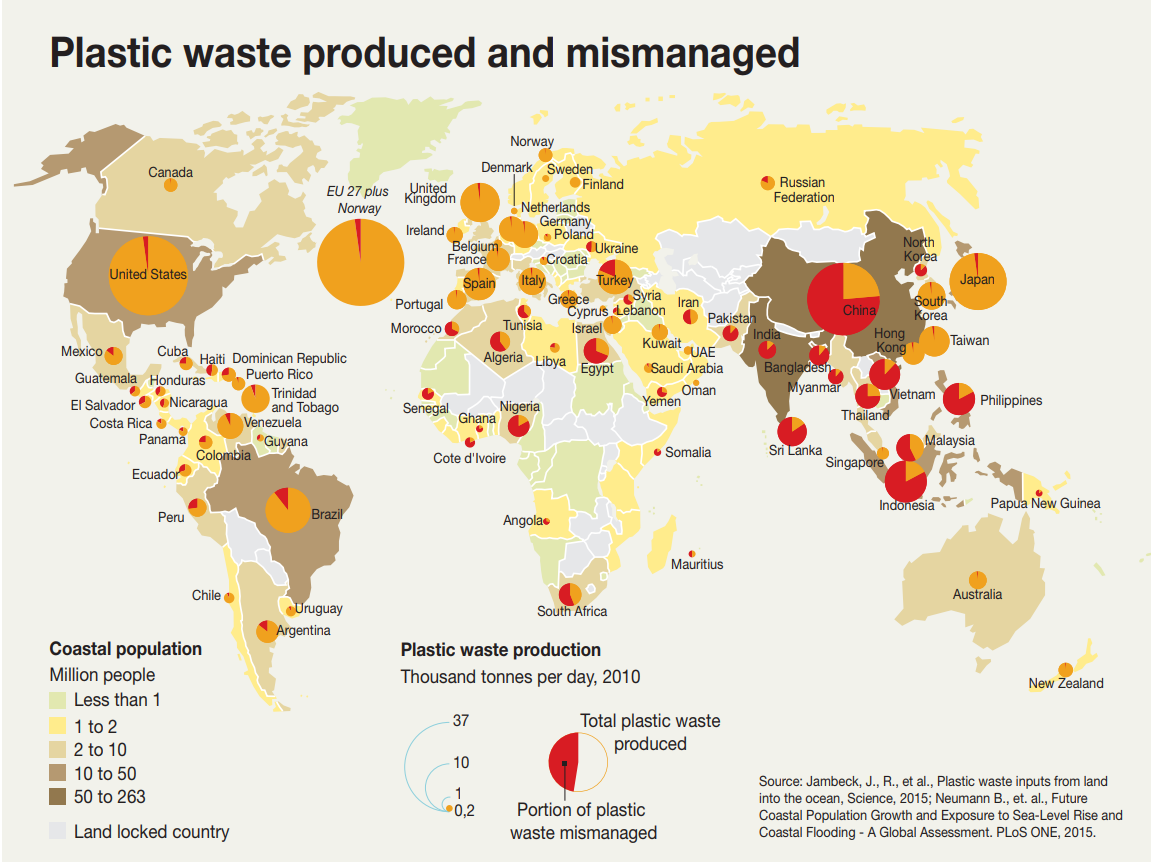}
  \caption{This figures shows global plastic wasted production levels and mismanagement \cite{2016UNEP-MarineLitterVitalGraphics}. Nation colors show the population coastal density with darker colors indicating higher densities. Pie charts shows plastic waste production in 2010 using units of thousands tonnes per day. The red coloring indicates the portion of mismanaged waste.}
  \label{fig:mismanagement}
\end{figure}

Plastic fragmentation is a key problem derived from plastic pollution in waterways. In Kershaw \et United Nations Environmental Program (UNEP) report \cite{2015UNEP-Kershaw-BiodegPlastic_MarineLitter,2016UNEP-MarineLitterVitalGraphics}, synthetic polymers though having widespread benefits also posed a great challenge. The challenge results from plastic's durability. While durability affords advantages in food preservation, electrical safety, etc., the poor management of post-use plastic along with plastic's durability becomes a significant problem in mitigating its impact
on the environment. Fig. \ref{fig:mismanagement} shows global plastic waste production levels per country as well as the proportion of mismanaged waste.
Plastics ubiquity in oceans results from poor waste management, failure to appreciate the value of ‘unwanted’ plastics, the underuse of market-based instruments (MBIs), and a lack of concern for the consequences \cite{2015UNEP-Kershaw-BiodegPlastic_MarineLitter,2016UNEP-MarineLitterVitalGraphics}. plastic's resistant nature to biodegratation leads to the exponential accumulation of the plastic waste mass in the ocean leading to physical and chemical risks to the marine environment. 
Even if biodegradable plastic was used, such plastics require typical industrial composting conditions---prolonged temperatures above 50$\degree$ C---to be completely broken down. Conditions that may never occur in the oceanic environment.

Non-biodegradable polymers (like polyethylene) weather and fragment in response to UV radiation, thermal oxidation, and microbial activity \cite{2015UNEP-Kershaw-BiodegPlastic_MarineLitter}. UV radiation, the the dominant agent, causes embrittlement, cracking and fragmentation, leading to the microplastic production  \cite{2011MPB-Andradi-MicroPlastics}. Some polymers are manufactured with a metal-based additive that results in even faster fragmentation. Plastic fragmentation then dissipates small plastic parts (microplastics) into the water.

Microplastics, (plastic debris less than 5 mm in diameter) have become ubiquitous in oceans. It is hard to exactly assess its quantity due to the small size of the particles and the fact that little is known about the chemical reactions and the extent of its incorporation into the trophic chain. However, numerous investigations are being conducted into the implications of organisms’ exposure to and intake of plastic particles. With this limited knowledge of the ultimate ecological effects of microplastics and nanoplastics, there are many concerns over potential deleterious effects to affected  ecosystems \cite{2016UNEP-MarineLitterVitalGraphics}.
\subsection{Garbage Gyres}
\begin{figure}[t]
  \centering
  \includegraphics[width=\linewidth]{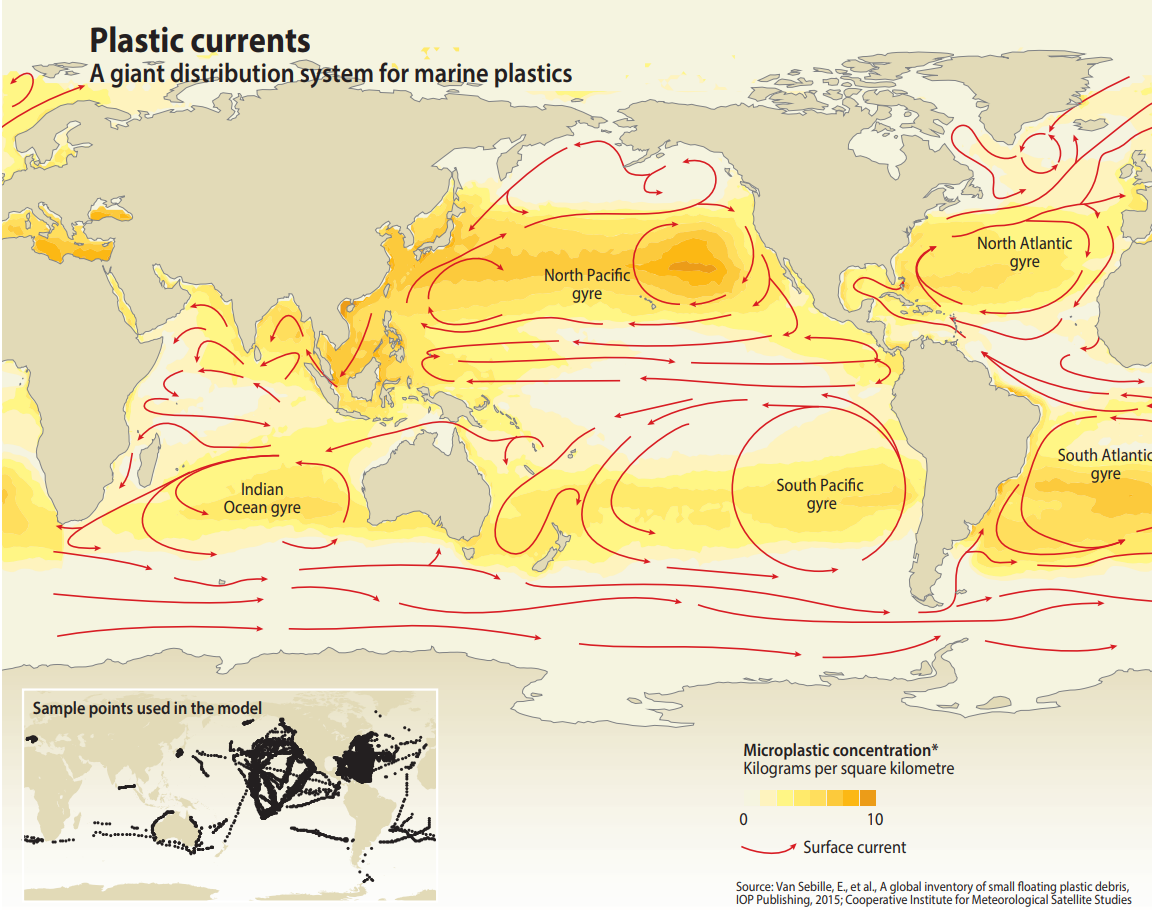}
  \caption{Illustration of garbage gyre patches around the world \cite{2016UNEP-MarineLitterVitalGraphics}.}
  \label{fig:gyres}
\end{figure}

We now turn our attention to the plastic debris that remains. 
Floating plastic debris accumulates in large patch regions called gyres as a result of ocean current patterns. The largest one is called the Great Pacific Garbage Patch (GPGP) \cite{2018SR-lebreton-GreatPacificGarbagePatch}. Fig. \ref{fig:gyres} shows the five main gyres around the world. 
Lebreton \et, provide evidence that the GPGP is rapidly accumulating plastic \cite{2018SR-lebreton-GreatPacificGarbagePatch}. The patch, equivalent to twice the area of Texas, changes shape and location easily due to seasonal and interannual variabilities of winds and currents. 
It is estimated that the patch contains between 1.1 and 3.6 trillion pieces of plastic and at least 79,000 tonnes when considering the central more dense portion of the patch. This is 4-16 times higher that previously calculated. \cite{2018SR-lebreton-GreatPacificGarbagePatch}. 
The patch's waste concentration densities vary concentrically. The center concentration levels reach 100s of kg/km$^2$, while decreasing down to 10 kg/km$^2$ in the outermost region. This densely distributed patch thus scatters and does not form a solid mass.
Within the patch, 92\% of the debris found in the patch consists of objects larger than 0.5 cm, and three quarters of the total mass is made of macro- and megaplastics (objects larger than 0.5m). However, in terms of object count, 94\% of the total is represented by microplastics. \cite{2018SR-lebreton-GreatPacificGarbagePatch}. Fishing nets also notably occupied around 46\% of the patch. 
Plastics tend to persist in the GPGP and frequencly fragment and disperse. Microplastics are found both on the surface and in the underlying water column as far as the ocean floor.  
\subsection{Marine Life Damage}
Recently, there has been much publicity about stomach contents of dead seabirds, whales, dolphins, and turtles caught in floating debris or wearing discarded plastic rubbish. By 2050, 99\% of seabirds are estimated to have had ingested plastic and currently that litter harms over 600 marine species \cite{2018WorldEnvironmentDay-UN}. However, small organisms living both on the seabed and in the water column are also suffering \cite{2016UNEP-MarineLitterVitalGraphics}. A study by Rochman \et \cite{2015SR-Rochman-AnthropgenicDebris} states that anthropogenic marine debris is observed throughout the ocean: from beaches and shallow coral reefs to the deep sea. Plastic particles have been found in hundreds of species of marine organisms, including many species of fish and shellfish sold for human consumption. The study found plastic in 25\% of the fish purchased from markets in the United States and Indonesia. Furthermore, Kuhn \et revealed that out of the examined species, marine litter was found in 100\% of turtles, 59\% of whales, 36\% of seals, and 40\% of seabird species. 
\begin{figure}[t]
  \centering
  \includegraphics[width=\linewidth]{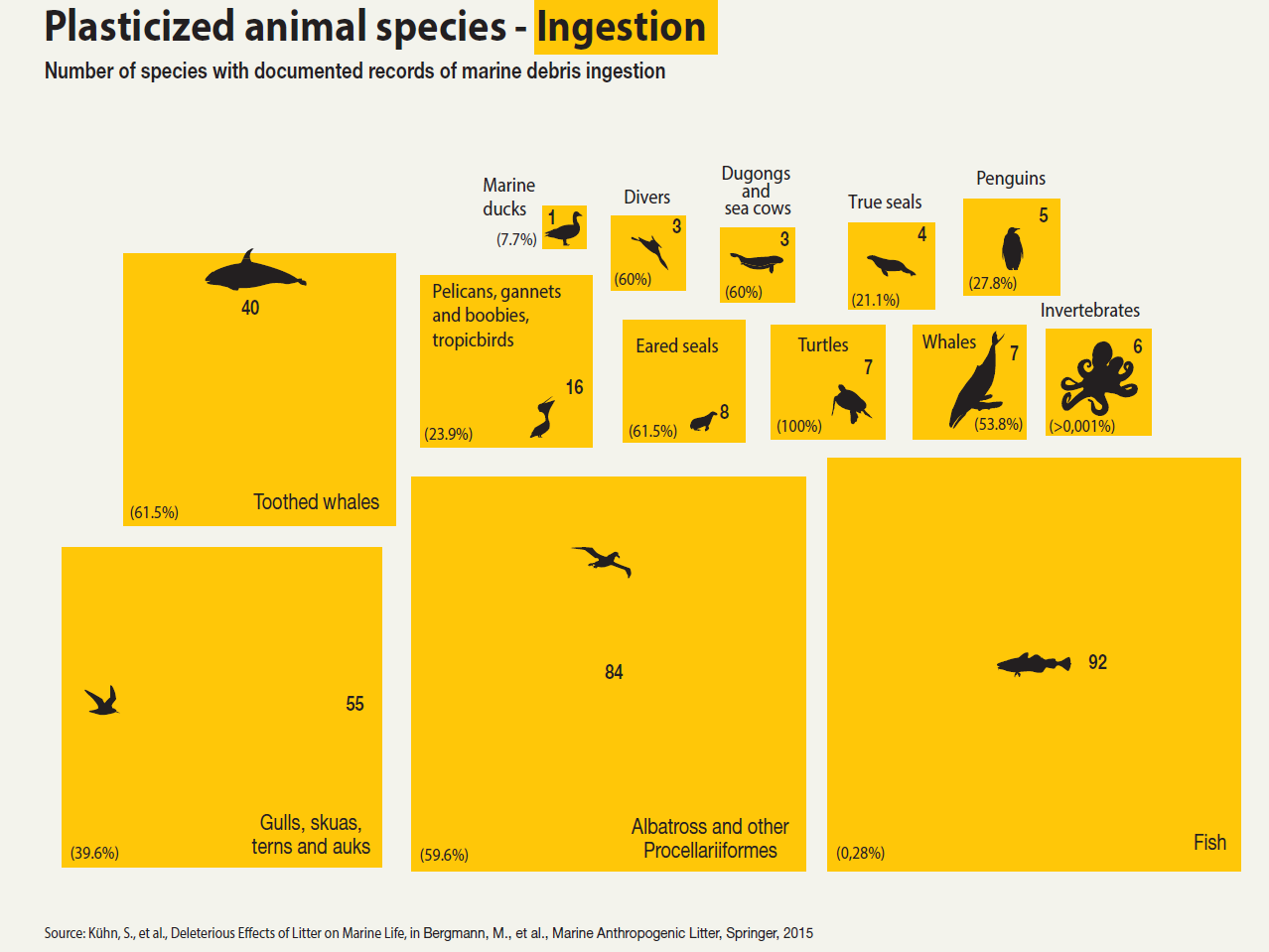}
  \caption{Kuhn \et revealed that out of the examined species, marine litter was found in 100\% of turtles, 59\% of whales, 36\% of seals, and 40\% of seabird species \cite{2015Springer-Huhn-DeleteriousEffects_Litter}.}
  \label{fig:ingested_plastic}
\end{figure}
In addition to the physical redistribution of plastics linked to wind and waves, there are a host of other process that redistribute plastics. One such process is the redistribution caused by organisms or animals. Microplastics are consumed by plankton, which is consumed by sediment ingesting organisms or small fish, which in turn are consumed by larger fish and so on \cite{2013EST-Cole-MicroplasticIngestion}. Birds and other mammals also play a role in the redistribution which exacerbates damage to marine life \cite{2016UNEP-MarineLitterVitalGraphics}.

\subsection{Summary}
In summary, a large number of challenges exist: poor waste management (especially in poor nations) leads to polluted oceans: from coastal regions, to open ocean waters, to underwater pollution. Plastic's durability leads to its persistent presence in the waters. The accumulation of macro and mega plastics have acquired dimensions and weights that are hard to conceive. Both macro and microplastics cause damage to marine animals and ecosystems in complex ways ranging from physical to biological processes. That damage also reaches humans when poisoned animals are consumed.
\section{Robot Trash Collection}\label{sec:robot_trash_collection}
This section surveys trash collecting robots. While the author sought to investigate maritime trash collection, the search only yielded works focused on the land domain. 

In \cite{1995AI-Balck-MultiagentBotTrashCollecTeam}, a team of cooperating robots worked to clean an office environment as quickly as possible within 10 minutes. Waste consisted of wads of paper, Styrofoam, coffee cups, and soda cans. Robots needed to collect waste and place it in trash cans while clearing tables, chairs, and desks. The design is now archaic. 
In \cite{1997RAS-nolfi-GarbCollecBot}, genetic evolution was used to implement a control system that performed a non-trivial sequence of behaviors when properly constrained in a simple simulated environment. In particular, a mobile robot was able to navigate and grasp simulated trash objects and return them to a collection area. 
In \cite{1998ICRA-Brumitt-Grammps}, used a general-purpose interpreted grammar for task definition along with dynamic planning techniques. The approach works on a general class of navigation systems and heterogeneous robots providing optimal execution given current world knowledge. The system worked on a simple simulation and real-world environment that consisted of two robots.
In \cite{2003EC-Vargas-ImmunoGenetNet}, used an immuno-genetic network for autonomous navigation. The algorithm combined an evolutionary algorithm with a continuous immune network model. The immuno-genetic network acted as a decision making process whilst the evolutionary algorithm defined the network structure. The algorithms were trained in simulation and tested with altered environments. Preliminary experiments were also done on a mobile robot in a simplified environment. 

In \cite{2009Dustbot}, the DustBot project initiative was launched to promote the design, development, and testing of systems for urban hygiene based on autonomous and cooperating robots. Improvements in robot localization and navigation (including obstacle avoidance) were obtained. Global (high-level) controllers yielded obstacle-free paths at the environment level, whilst sensors achieved local reactive navigation crucial in non-static environments. Realtime solutions used an various camera types, infrared, microwave, laser, and ultrasound. An equivalent of such project would be extremely helpful to incentivize research in water pollution collection. The scope of the problem for water-based pollution (rivers, coastlines, free ocean waters, underwater pollution, and macro- and micro-scales) is much larger and more complex. Energy requirements, collection quantities, size, and weight are much grander. Possible information collection consists of a much broader set of sources (satellite data, local sensors, networks of agents). Additionally, the environment in which trash collection takes place--the ocean--consistently introduces dynamics that alter the environmental map making this a much harder problem. 

In \cite{2011ICRA-Ferri-Dustcart}, a dustcart robot was deployed to collect garbage from homes and deposit it in collection points in a small urban town in Italy. The system employed optical beacons and localization heuristics to navigate GPS-denied environments. This work is a good example of more recent efforts in deploying research work in real human environments. 

In \cite{2010HRI-Yamaji-SocialTrashBot}, a social trash bot was assisted by children to help collect trash through social rapport. The robot required human assistance in the form of interactive behaviors and vocalizations to accomplish its goals. Perhaps, such work could be extended to enhance already existing human efforts to clean river and ocean waters. It would also be interesting to explore the possibility of leveraging marine animals to assist with cleanup if it were possible to influence their behavior (without risk to their health) and couple their behavior with tools for garbage collection. 
Developments as the ones listed above might be useful to cleanup beaches. The literature shows some early prototypes in \cite{how2002biologically,wattanasophon2014garbage}. A number of challenges remain before these robots can be deployed. Particularly its interaction dynamics in beaches with human crowds, deployment and maintenance in poorer nations, and it's overall effectiveness. 
\section{Plastic Waste Collection} \label{sec:surface_collection}
\subsection{Ocean Collection}
\subsubsection{Manta by SeaCleaners}
SeaCleaners is a French organization that is seeking to tackle macroplastic maritime pollution \cite{2018SeaCleaners}. Their goal is to work near the source of macro plastic waste pollution so as to collect large quantities of plastic before it fragments and dissipates. 
To this end, they are undertaking the design of their Manta vessel. It is designed to be a large vehicle with state-of-the-art technologies in clean energy production, garbage collection, and handling. 
The boat is set to consist with 2000 m$^2$ of solar panels along with 100 tonnes of batteries, a hybrid propulsion system (it has both sail and electric motors) providing the MANTA with autonomy for travel and collection phases. 
The Manta is to rapidly intervene in the most polluted areas; whether in deep waters, coastlines, or in the estuaries of the ten great rivers from which 90\% of all plastic waste reaches our oceans. Manta consists of 2 Darrieus wind-turbines – producing 500 Kw/h.

Three collecting treadmills are to be installed between the hulls of the ship to pull up big quantities of plastics that are manually sorted. The plastic waste is transported by a conveyor system to the manual sorting area where the reusable plastic waste is separated from other debris and compacted into 1m$^3$ bales. Upto 250 tonnes of waste in its hulls before repatriating them on the land, where they will be supported by adapted recycling centers. Additionally, through the use of two cranes, MANTA can also remove larger floating debris from the water, such as nets or containers lost by other vessels or debris from the land

The MANTA will also be equipped with a complete scientific laboratory which will enable the waste to be studied in terms of geolocation, quantity and quality that will be shared openly with the international community. During collection, the Manta's audio emission system will divert fish from its route so as to prevent them from being captured accidentally. 

It seems that SeaCleaners has a very strong commercial proposition. It integrates innovative technologies to achieve energy autonomy, semi-autonomous navigation, as well as surface and shallow depth collection of material. Though the 250 tonne collection specification per mission is minuscule when compared to the current estimated amount of 322 million tonnes, the Manta vessel could make a sizeable contribution if it runs efficiently over the long term and if the number of vessels were to increase. A strong financial backing would be necessary for scaling to be possible; it is to be seen if this would be best achieved through government, international organizations, private funding, or through a combined program. 
\subsubsection{Ranmarine Technology}
Ranmarine is an international company that has recently introduced a light-weight vessel inspired by whale sharks called Wasteshark \cite{2018RanMarine}. As the whale shark, it is designed to swim through water while eating its prey with minimum effort and maximum efficiency. This vessel weighs 39kg and measures 1.5m long by 1.1m wide by 0.45m deep. It has a thrust of 5.1 kgf and it can carry 200 litres of waste. The vessel is designed to be efficient, long-lived, non-threatening, and unobtrusive (specified data was not found in ). Wasteshark also performs data collection and transmission duties. Its sensory equipment include depth, temperature, and water quality tracking 16 hours per day. As for collection capabilities, plastics and micro-plastics, alien vegetation (like duckweed), and floating debris can be collected. As for navigation capabilities, the vessel can be remote-controlled or follow pre-established navigation waypoints and has zero greenhouse emissions.

Wasteshark might be a good solution in rivers and coastlines that are mildly polluted. Perhaps, if one could scale to hundreds, these vessels could form a multiagent team with heavy duty vessels like Manta to clean up outstanding debris. Goal optimization in multiagent teams has been intensively studied in the last few decades and powerful solutions exist to achieve complex goals. 
\subsection{River Collection}
Only one example of river cleaning robots was found in our search. Urban Rivers is a river cleaning robot project, funded in Kickstarter\footnote{www.kickstarter.com}, and currently working on their alpha release \cite{2018UrbanRivers}.

The Urban River's robot is designed to herd trash to a safe location in the Chicago River in the United States. Interestingly, they decided to use crowd sourcing to control the robot and in this way letting the public drive the robot and clean the river.  

The main challenge for this design is to reduce control latency at all levels. Communication latency was solved by installing a gigabit internet connection in an antenna with line of sight to the portion of the river where collection would take place. The antenna used is an Ubiquiti Sector antenna along with a lightbeam antenna. Motors can be controlled in real time through an android app. Different propulsion methods are still under trial. Video is streamed using H264 encoded streams. In terms of trash collection, trials with a bucket system or a ``belt and belly'' system are currently underway. The bucket system seems a discrete-time catching mechanism that would pick up trash seen in front of the robot. The ``belt and belly'' system consists of a rotating belt with hooks that would pull the trash from the water unto the vessel. When the trash reaches the upper edge of the band, the trash would then fall unto a container for collection.

One of the risks mentioned by the team is vandalism and software security. Vandalism may be mitigated through the use of GPS tracking and public camera surveillance. There is also the risk of hacking. Efforts have been undertaken to install a safety tether into the robot and a GPS cage that would limit the locations the robot can visit and in this way reduce the risk of unwanted collisions.

There is still no reporting on the waste collection capacity this robot might have for herding surface waste. It's also unclear if the performance would vary in different areas of the river (a study of the river itself might be useful to understand its dynamics). It's also unclear if this kind of technology would work (or even be desirable) in poorer nations (perhaps economics would dictate that people perform this job instead of a robot).
\subsection{Non-Robotic Efforts}
Ocean Cleanup is designed by an International foundation that designed a surface and near-surface collection system that has been in development for the last five years and one that will be undergoing real-ocean trials on September 2018 \cite{2018OceanCleanUp}. The system consists of a 600-meter-long floater that sits on the surface of the water with a tapered 3-meter-deep skirt attached below. The floater provides buoyancy to the system and prevents plastic from flowing over it, while the skirt stops debris from escaping underneath.
Both the plastic waste and the system are carried by currents. However, wind and waves propel only the system as the floater sits just above the water surface, while the plastic is primarily (just) beneath it. As a result, the system moves faster than the plastic, allowing the system to capture the waste. Furthermore, the system's U-shape, leads plastic to be collected at the center of the system. 
Ocean Cleanup is designed to capture plastics ranging from small pieces just millimeters in size, up to large debris, including massive discarded fishing nets (they can span tens of meters). 
After capturing the waste, the Ocean Cleanup plan states that a vessel acting as a garbage truck would remove collected garbage every few months and plastic would finally processed on land and sorted for recycling. No mention was made of the type of vessel that would perform the collection. Perhaps the Manta vessel or a similar model could work well in performing these joint functions.  

According to reports from Ocean Cleanup, their simulation models predict that a full-scale system roll-out (a fleet of approximately 60 systems) could clean 50\% of the GPGP in just five years and 90\% of ocean plastic by 2040. This prediction is of extreme significance. If successful, the impact of this solution could not be understated.
\subsection{Underwater Collection} \label{subsec:underwater}
Trash presence underwater or buried under the marine floor is similar a huge problem. From the UNEPs study in \cite{2016UNEP-MarineLitterVitalGraphics}, if one considers all plastics produced since 1950, the open ocean waters have 39\% of the total plastic trash spiraling through its waters, or around 34 million tonnes! Similarly, the coastline and seafloor may account for 33.7\% of plastic waste or around 29 million tonnes. What systems could possibly meet this challenge?

Two patents seem to try to address this issue. First in \cite{irene2014trash}, a low-depth mechanism is described. The system enables divers to remove debris from water through a mechanical line attached to a boat and extending down into the water. Hooks or shelf nets are mechanically coupled to the line and configured to receive debris from a diver. Then in \cite{stilwagen2017trash}, trash can be removed from a water bottom by the use of sled. The sled uses a jetting aeration system to agitate debris and a vacuum to suck in the trash. A conveyor belt can be used to transport trash to the surface. 

Other possibilities might include the use of underwater manipulators to perform the individual picking of waste items. According to \cite{2016RAM-Khatib-OceanOne}, recent advances in underwater manipulators include Schilling’s ATLAS 7R hydraulic arms, the semi-autonomous vehicle for intervention missions (SAUVIM) MARIS 7080 electrical arms, ECA Group’s ARM 5E lightweight arm, the SAUVIM, and the reconfigurable autonomous underwater vehicle for intervention (RAUVI). Another possibility might include oceanic avatars. ``Ocean One'' is a deep sea water human avatar that boasts fine manipulation skills underwater as well as allowing human users to control the robot thus leveraging human's intuition and expertise and cognitive abilities \cite{2016RAM-Khatib-OceanOne,2017IJRR-Stuart-OceanOne}. The avatar provides a high degree of autonomy in physical interaction but is also connected to human experts through an intuitive interface. Perhaps similar technology could be used to pick trash in the water but also that which has been slightly buried under the sea floor. The robot 
\section{Global Campaigns}\label{sec:global_campaigns}
The plastic waste catastrophe is beginning to catch serious attention worldwide. The United Nations, many countries, international organizations, companies, and grassroots efforts have began to mobilize themselves in important ways. The UN recently started their ``Clean Seas Initiative'' \cite{2018Cleanseas-UN}. As of March 2018, 40 nations have committed to work with the UNs Environmental committee to introduce regulations and incentives to tackle marine litter, promote public awareness and, exchange best practices. India, Rwanda, and Bangladesh have or will soon ban single-use plastic bags. France will ban single-use plastic cups, plates and cutlery by 2020. The US and the UK banned microbeads in cosmetics in 2017 and Canada has added them to its list of toxic substances. 

Several other promising international policy initiatives have emerged recently. The Sustainable Development Goals (in particular Goal 12, which seeks to ensure sustainable consumption and production patterns, and Goal 14, which seeks to conserve the oceans and marine resources), the G7 Alliance on Resource Efficiency, and the G20 Action Plan on Marine Litter \cite{2018MarinePolicy-Penca-EuropPlasticStrat}. The European Union also launched its Plastic's strategy policy on January, 2018. Beyond simply setting regulatory measures on industries regarding waste leakage into natural systems, the policy consists of a circular focus which address transformative processes across supply chains and is sure to have important positive effects in curtailing the problem.
\section{Discussion and Conclusion} \label{sec:conclusion}
This paper presented an overview of the main problems posed by plastic waste in rivers and oceans as well as how robotic research and robotic projects address these issues. Plastic waste is a world catastrophe that poses enormous challenges. Very little to none is being done in the way of robotic research to find solutions. Most work is coming from individual companies and foundations. 

Given current progress, it seems that robotics can play a complimentary but important role in cleaning up our oceans and rivers. To date, the Ocean Cleanup project stands as the best solution for surface waste collection \cite{2018OceanCleanUp} so far. This effort requires of marine vessels to pick and dispose collected trash. Larger robotic vessels like the Manta and the smaller vessel like the Wasteshark could collaborate in multiagent teams to complete such tasks. With regards to underwater collection, no serious effort has been identified. This paper offered some possibilities based on existing underwater rover technologies as described in Sec. \ref{subsec:underwater}. Finally, the collection of microplastics seems even more challenging; leaving us with only questions at this moment.

As seen in Sec. \ref{sec:global_campaigns}, policy developments around the world offer hope to curtail waste from the source. Nonetheless, considerable time will pass before significant effects are felt by such policies. Even so, it has been seen that at times nations turn away from their commitments \footnote{This BBC report details how many countries have failed in their commitments (https://www.bbc.com/news/science-environment-44359614). Published 6 June 2018 and accessed September 2018.}. Furthermore, the longer plastics remain in the ocean, the greater the maritime environment will be damaged. For this reason, it is imperative that no effort is spared in the collection process. While there is a significant progress being made in basic research for surface and underwater autonomous robot systems; to date, no robotics applied research has focused on solving the various problems posed by plastic water waste. This paper's goal is to raise awareness and spur interest in this pressing topic.
\section{Disclosure}
The author is not a member, nor does he receive money, funding, or any other benefit from all of the cited works or referenced organizations in this paper. The main goal is to inform on the current state of affairs of robotic technology in dealing with current marine littering from plastics.
\bibliographystyle{IEEEtran}
\bibliography{IEEEabrv,Xbib}
\end{document}